# Counter-Inferential Behavior in Natural and Artificial Cognitive Systems

Serge Dolgikh[1[0000-0001-5929-8954]]
Dept. of Artificial Intelligence,
Lviv Polytechnic National University
Lviv, Ukraine

**Abstract.** This study explores the emergence of counter-inferential behavior in natural and artificial cognitive systems, that is, patterns in which agents misattribute empirical success or suppress adaptation, leading to epistemic rigidity or maladaptive stability. We analyze archetypal scenarios in which such behavior arises: reinforcement of stability through reward imbalance, meta-cognitive attribution of success to internal superiority, and protective reframing under perceived model fragility. Rather than arising from noise or flawed design, these behaviors emerge through structured interactions between internal information models, empirical feedback, and higher-order evaluation mechanisms. Drawing on evidence from artificial systems, biological cognition, human psychology, and social dynamics, we identify counter-inferential behavior as a general cognitive vulnerability that can manifest even in otherwise well-adapted systems. The findings highlight the importance of preserving minimal adaptive activation under stable conditions and suggest design principles for cognitive architectures that can resist rigidity under informational stress.

## 1 Introduction

The balance between two fundamental aspects and objectives of learning: maintaining the stability of the knowledge base and updating it with new information has been observed and examined in various areas related to the general domain of cognition. These areas include biological cognition (including animal cognition), human intelligence, theoretical and empirical cognitive science, machine intelligence, and artificial intelligent agents.

In this work, we attempt to address the question: can advanced intelligent systems, whether biological or artificial, that attain the capacity to construct sophisticated and fluent information models of their sensory environments develop a persistent preference for inner stability over ongoing relevance and attunement to the environment? If so, under what conditions does this occur, particularly from the perspective of the interaction of cognitive functions and modules, information dynamics within the system, and its cognitive functions and components?

While heuristic- or error-driven suppression of information has been widely documented, the phenomenon of strategic, intentional, and model-aware suppression, wherein an intelligent system attempts to maintain cognitive stability by deliberately and actively limiting new information intake remains relatively underexplored. In this work, we aim to investigate the underlying informational stimuli and cognitive drivers that promote such behaviors, with a focus on how reward balancing and internal model dynamics shape these choices in advanced autonomous cognitive systems, both natural and artificial.

It needs to be noted that in the context of this work, we use the terms counter-inferential and counter-Bayesian behavior or bias to describe deviations from expected or normatively rational inference, especially in systems with adaptive or meta-cognitive capabilities. These deviations may manifest internally as disbalanced evaluation or reward reassignment (bias) or externally rigid or unexpectedly conservative responses (behavior). While they differ slightly in emphasis, we will use the two terms largely interchangeably to avoid repetition and to accommodate both internal and external dimensions of the observed phenomena.

## 2 Related Work

Cognitive systems that operate in complex, mutable and volatile sensory environments must balance stability (coherence, robustness, learned patterns, knowledge and beliefs) with adaptability (plasticity, flexibility, fluency) of the information model. This tension appears as a well-known stability–plasticity dilemma in both neuroscience and the research in artificial intelligent systems. For example, a neural network must learn new inputs (plasticity) without catastrophically forgetting old memories (stability) [1].

The dilemma manifests itself as a balance or trade-off: if plasticity is too high, new learning can risk erasing prior knowledge ("catastrophic forgetting", [1,2]); if stability is too high, the model may not be able to encode new data efficiently.

Other phenomena illustrate this continuum from the opposite end of the balance. For instance, the entrenchment effect refers to early-learned information becoming overly durable, making later learning harder. Connectionist simulations show that items learned first are "entrenched" in the network weights and produce fewer errors even after training frequency is controlled [1,3].

This reflects the same stability–plasticity balance: the brain's early-acquired knowledge tends to be highly stable (robust), at the cost of reduced flexibility for incorporating later information. The problem can be summarized as the Stability–Plasticity dilemma: balancing integration of new data (plasticity) against retention of old knowledge (stability). Too much plasticity → catastrophic forgetting; too much stability → learning rigid, inflexible models [1]. It is also known as the Stability–Flexibility (cognitive) trade-off: for example, in executive control, stability (focused maintenance of a task set) and flexibility (the capacity to accept and switch tasks) are often seen as opposite ends of a continuum [4].

In a simplest case, a single meta-control parameter, the updating threshold may govern this balance: lowering the threshold yields fast switches (high flexibility) but poor protection from distraction (low stability) [5] whereas biological intelligent systems combine complementary learning strategies that strive to optimize the balance between the objectives. For example, a structural solution where hippocampal (rapid) learning and cortical (slow) learning work together to balance plasticity and stability [1,6,7].

Closely related concepts have been actively researched in the field of artificial cognitive systems: meta-control parameters such as the updating threshold [8] or the learning rate in probabilistic models tunes the trade-off. A high learning rate (corresponds to low threshold) integrates new evidence quickly (adaptable model) but prone to overwriting prior knowledge (less unstable), whereas a low learning rate preserves older beliefs (stable) at the cost of slow adaptation (less fluent/adaptable).

In predictive-processing models, the precision of prior beliefs encodes stability. High-precision priors strongly bias perception and learning, making the agent "inertial" or "resilient as inertia" [9]. Such high precision yields robust models that resist change, while low precision makes beliefs more malleable. For instance, a person with overly precise priors (e.g. a flat-Earth believer) may discount disconfirming evidence effecting a highly resilient (stable) information model however at a cost of accuracy and fluency or even a substantial deviation from the empirical reality.

These studies resonate with bias–variance trade-offs in machine learning models. A "stable" model (high bias, low variance) generalizes across noise but cannot fit peculiarities of data, while a highly flexible model (low bias, high variance) fits recent data well but may forget or discount patterns and trends learned earlier [4].

Research in cognitive control explicitly frames this trade-off. Cognitive stability means maintaining a task set or belief in the face of interference, whereas cognitive flexibility means rapidly shifting to new tasks or updating beliefs when the environment changes [5]. Traditionally it was thought that stability and flexibility were opposed: more of one necessarily means less of the other but recent work shows that they can be independently regulated [10].

Likewise, exploration–exploitation trade-offs in decision-making echo this balance. Focusing on known, rewarding actions (exploitation) preserves current knowledge, whereas exploring new actions introduces novel inputs and broadens future adaptability. Indeed, investing resources in exploration "may contribute to increase the future flexibility of the system by broadening the range of inputs" it can handle [4].

Active-inference (Bayesian brain) frameworks cast stability versus adaptability in terms of precision weighting of prediction errors. Strong (high-precision) priors lead to "inertia": they act as powerful inductive biases that resist updating. Such agents are model-coherent but can be inflexible as they tend to set firmly in their current beliefs and ignore even robust counterevidence. Conversely, low-precision priors allow rapid updating and sensitivity to new data. Thus, resilience as inertia (structural robustness) trades off against plasticity (learning new contingencies) [9,11].

An extreme consequence of this imbalance is counter-Bayesian or counter-normative updating. For example, perceptual experiments show that people sometimes interpret ambiguous sensory data in the direction opposite to their priors. For example, in the classic size – weight illusion, one can be expected (normatively) to experience the smaller and larger objects as equally heavy given equal weight, but subjects actually perceive the smaller object as heavier [12]. Such patterns can arise when the system's stability mechanisms (strong expectations or priors) conflict with new evidence in a way that reverses the normal update rule. In effect, excessive stability or entrenched heuristic processes can lead to "perverse" updates that defy Bayesian norms.

While the prior literature has extensively documented various manifestations of cognitive rigidity, bias, and stability–plasticity trade-offs across biological, psychological, and artificial systems, much of this work has focused on behavioral outcomes or specific domain-level mechanisms. These studies have provided valuable insight into what forms counter-adaptive, counter-inferential, "anti-Bayesian" behavior may take and where it appears, yet they often stop short of addressing the deeper information dynamics that drive such behaviors across systems of varying complexity.

The present study builds upon this foundation by proposing a general, system-level account of counter-inferential behavior grounded in the trade-off between adaptability and stability rewards within cognitive architectures. Rather than treating bias or rigidity as a flaw or malfunction, it is approached here as a strategic or emergent outcome of reward-mediated information model regulation. This information-dynamical aspect enables a unified treatment of diverse scenarios spanning biological cognition, reflective human behavior, and adaptive artificial systems by emphasizing how feedback, internal model evaluation, and reward balance shape cognitive evolution over time. In doing so, the study seeks to move beyond domain-specific descriptions toward identifying possible common mechanisms and structural vulnerabilities inherent in adaptive cognitive systems.

The rest of this paper is organized as follows. Section 3 introduces and discusses the concepts of cognitive autonomy and the balance of cognitive rewards, with a focus on how autonomous cognitive systems process information derived from sensory interaction with their environment. Section 4 analyzes specific information-dynamic scenarios that give rise to counter-inferential biases and behaviors. Section 5 reviews empirical findings across several domains that support and contextualize the theoretical framework. Sections 6 provides a synthesis of the results and a concluding summary of the study.

## 3 Cognitive Autonomy and the Stability-Adaptability Balance

### 3.1 Cognitive Autonomy and Autonomous Sustainable Information Models

We will be dealing with intelligent systems, biological or artificial, that attained or are capable of attaining the state of cognitive autonomy: the ability to develop and sustain sophisticated, comprehensive, accurate and fluent information models of their sensory environments.

These information models have fine-grained predictive and explanatory power, the accuracy and temporal stability/sustainability being comprehensive, coherent and self-reinforcing so that they can persist over certain and in some cases, extended periods of time without defining and direct dependence on empirical updates.

We will refer to models in this class as Autonomous Sustainable Information Models (ASIM) or self-sustainable information model.

The key characteristics of these models can be outlined as:

- **Descriptive power**, comprehension, granularity: sufficiently rich informative and discriminative description of the sensory domain or part(s) thereof
- **Coherence**: models are logically and cognitively sound and cohesive, without essential and critical self-contradictions.
- **Sustainability**, stability: the model can persist in the cognitive space of the learner for extended periods of time with minimal direct environmental feedback.

Models, as defined, are characterized by their ability to persist independently of empirical inputs. However, to maintain adequacy and fluency in relation to the actual state of the external environment, they rely on mechanisms that balance the stability of the model with its adaptability to new information. As discussed earlier, this balance between stability and adaptability is crucial for ensuring that the model remains responsive to environmental changes while maintaining internal consistency.

### 3.2 The Balance of Cognitive Rewards

We assume that at any given cognitive state, condition, or sensory stimulus, the system's preference and decision: whether to engage in exploration, acquire new knowledge, and update its existing informational base and structures, or conversely to preserve and safeguard them can be effectively modeled as a balance between two factors, cognitive rewards. This does not imply that the underlying mechanism must operate explicitly through weighted reward/penalty signals; rather, we posit that its behavior can be approximated with sufficient accuracy for our purposes by such a model framework.

Formally, cognitive rewards operate as functions (or, where needed, functionals) defined on the joint space of cognitive states and sensory inputs. Let $C$ denote the space of possible cognitive states of the system, and $S$, the space of sensory stimuli. Then, a cognitive reward can be defined as a function on the space of interactive states $(c, s) \in C \times S$; $R_C: C \times S \rightarrow \mathbb{R}$

$$R_C: C \times S \rightarrow \mathbb{R} \tag{1}$$

mapping an interactive state $(c,s)$ to a scalar reward value that influences the system's decision-making dynamics.

1. **Stability cognitive reward** $R_S$ determines the cognitive and empirical benefit of maintaining a stable, constant state of the information model. The factors that substantiate the benefit of maintaining a stable information model are:

- Operational and resource efficiency: reduces the need for frequent retraining/recalibration of functions/methods that depend on the model.
- Structural and operational coherence: Preserves the integration of various subsystems in interpretation of sensory stimuli and construction of the responses.

Stability incentive thus reduces or mitigates the impact of informational volatility: highly dynamic updates can force constant revision of dependencies, risking breakdowns in derived systems and action pipelines.

2. **Adaptability/fluency cognitive reward** $R_A$ incentivizes the drive to seek new information and update or improve the information model for close synchronization and rapport with the sensory environment. The cognitive benefits associated with this reward are:

- Predictive accuracy due to more precise information model of the sensory environment.
- Adaptability and resilience to significant changes in the environment.

Continuous adaptation and synchronization of information model and response-producing functions is essential for long-term survival/performance in non-stationary environments.

Next, we define a balance function $B$ that maps the pair of rewards ($R_S$, $R_A$) to a scalar value that reflects their joint influence on the system's decision-making. The function is designed to embody the principle of counteraction between the stability and adaptability rewards again, not necessarily in a direct, literal way. While a basic implementation might take the form $B(R_S, R_A) = R_S - R_A$ or a normalized ratio, we leave the specific form of $B$ open to accommodate nonlinearities, context dependencies, or functional compositions that better reflect the system's dynamics [10].

$$B(R_S, R_A) \to \mathbb{R} \tag{2}$$

### 3.3 The Information Dynamics Basis of Counter-Inferential Bias

In adaptive cognitive systems in the considered (ASIM) class, normal information behavior arises from a dynamic balance between two fundamental drives: adaptability and stability. This balance governs the internal reward structure guiding model updates and sensory engagement. When functioning optimally, it enables the system to construct and maintain an information model that is both accurate and resilient: i.e., capable of capturing relevant environmental patterns while remaining coherent and stable over time [3-5].

However, the inherent tension is always present between adaptability, updating of the information model that improves its fidelity but risks undermining internal consistency, while stability preserves coherence but may lead to loss of rapport with the sensory environment. An overemphasis on stability can result in cognitive stagnation where the system resists incorporating new evidence, even of substantial or even existential significance. Conversely, the bias to over-adaptation can cause the information model to become brittle, reacting excessively to insignificant transient patterns or statistical noise.

When the internal reward for stability outweighs the reward for empirical accuracy, the system may:

- Avoid or delay seeking new information.
- Filter or reinterpret evidence in ways that minimize disruptive updates.
- Prioritize internal coherence even at the cost of reduced responsiveness to real-world change.

Crucially, this pattern, referred to here as *counter-inferential bias*, *counter-Bayesian behavior* is not merely a malfunction or an error in reasoning. Rather, it reflects a rational trade-off enacted under constraints where the cost of frequent revision, instability, or uncertainty is perceived to outweigh the benefit of marginal gains in accuracy or adaptability.

In contrast, when adaptability is rewarded too heavily at the expense of stability, the system may become hypersensitive and fragile, continually restructuring its models and reducing its capacity to form lasting, generalizable knowledge.

Counter-inferential behaviors can thus emerge naturally from the underlying information dynamics when this balance is misaligned whether through design, misperception, or contextual pressure.

Maintaining an appropriate balance between these competing incentives is therefore vital: not only for successful adaptation to environmental signals, but mainly for the long-term coherence and viability of the cognitive system itself.

## 4 Information-Dynamic Mechanisms of Emergence of Counter-Inferential Bias

In this section we will consider several scenarios that can lead to the emergence and instatement of counter-Bayesian bias and behavior in intelligent systems, natural and artificial. Our analysis will follow a structured approach, examining the key drivers, mechanisms, facilitating conditions, and expected outcomes. This provides a comparative basis for understanding the diverse ways in which cognitive reward dynamics can bias information processing.

### 4.1 Success Saturation Mechanism

**Conditions of emergence**: this scenario can emerge in environments characterized by extended periods of sensory stability and consistent empirical success. The system, equipped with a sophisticated information model, observes a high degree of correlation between its predictions and external stimuli over time. The empirical success from these consistent predictions reinforces the model's current configuration.

**Key driver/cause**: repeated positive reinforcement of the system's current model stability, where sensory inputs remain predictable and consistent.

The underlying cause is an associative reinforcement loop where the system may interpret the coincidence of success and the current cognitive configuration as a deeper causal relationship, where the current state and stability of the model is perceived as the source of empirical success.

**Information dynamic mechanism**
The stability reward $R_S(c, s)$ can become increasingly dominant due to repeated empirical validation and reinforcement of the current information model, $M(\tau)$ while the adaptability reward $R_A(c, s)$ remains under-stimulated in environments with minimal novelty. Through reinforced correlation, the system may mistakenly learn or infer that the internal model's stability is causally responsible for the observed empirical success $E$. In reality, the true cause of success is the external environmental stability but the system interprets the correlation as evidence of its own internal efficacy, increasing the weighting of the stability reward over adaptability associating it with empirical success (Fig.1).

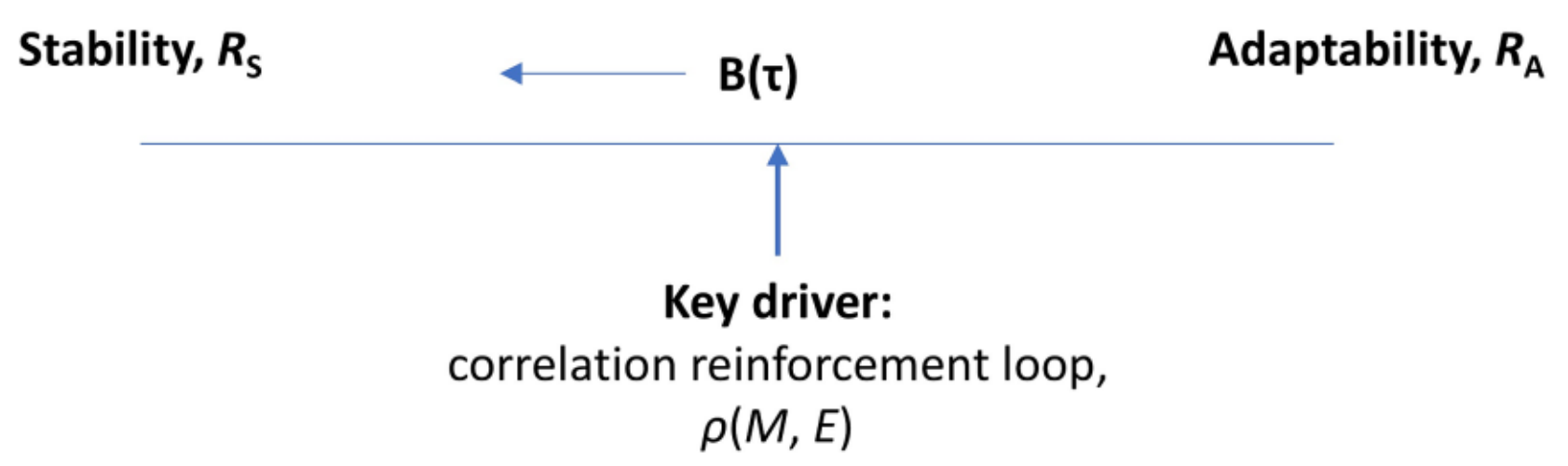

**Fig. 1.** Information dynamic mechanism, success saturation scenario.

A simple expression for the resulting shift of the reward balance function, the stability bias can be written as:

$$\Delta B(R_S, R_A) \rightarrow R_s \sim \eta\, S_f$$

where $S_f$: empirical success frequency/rate, $\eta$: a constant factor.

If unchecked, this dynamic mechanism can lead to a gradual shift in the reward balance, potentially incentivizing the system to move toward preservation and reducing its responsiveness to new information.

**Facilitating factors**: Several conditions can trigger or amplify the likelihood of this bias:

- Long-term environmental stability: a prolonged period of predictable and consistent sensory inputs.
- High initial model accuracy and precision: low frequency of significant prediction errors in the early stages, reinforcing stability.
- Structural/architectural coupling: a strong link between the internal model configuration and the feedback from empirical success, reinforcing the perceived efficacy of the model.

- Insufficient meta-cognitive oversight: lack of a higher-level system function to monitor the adaptability of the model, potentially leading to the oversight of risks associated with stasis.

**Potential outcomes**: As a result, the system may become overly conservative, favoring model stasis over adaptability. This can lead to cognitive inertia, where the system becomes increasingly resistant to change, even though the underlying mechanism was originally designed for adaptive intelligence.

When environmental shifts inevitably occur, the system may fail to adapt in a timely or effective manner. In extreme cases, this could result in catastrophic failure if the model is too rigid to accommodate new and unforeseen information.

**Interpretation, connections and implications**

This scenario highlights how reward imbalance can distort cognitive dynamics even without intentionality or error in design. The system, overemphasizing stability over adaptability, becomes increasingly resistant to change, which mirrors several well-documented phenomena:

- Overfitting to a stable regime: in both machine learning and cognitive systems, overfitting to consistent inputs can lead to a model that no longer generalizes well to new or varying data.
- Organizational complacency in human systems: long periods of stable success can lead to reduced vigilance and innovation, as organizations become comfortable with the status quo.
- Homeostasis-like behaviors in biology: biological systems can exhibit reduced responsiveness when in prolonged equilibrium, potentially compromising adaptability when environmental shifts occur.

This scenario reinforces the importance of maintaining a minimal activation of the adaptability channel $R_A$, even in the absence of obvious environmental change. This "exploration baseline" ensures the system remains flexible and capable of responding to unexpected shifts, avoiding overfitting to a state of stasis.

We will discuss empirical and research evidence, implications and connections of the information basis of counter-inferential biases and behaviors in greater detail in Section 5.

## 4.2 Overconfidence Bias

**Conditions of emergence**

This scenario can arise in environments characterized by prolonged periods of empirical success and stability, where the system does not merely correlate success with the stability of its internal model (as in Scenario 1), but elevates its interpretation to a meta-cognitive level. The system begins to attribute its consistent success to the intrinsic superiority or near-perfection of its cognitive model.

This represents a shift from straightforward associative learning: "my stable model correlates with success" to a higher-level inference: "my cognitive model is inherently superior and thus guarantees success."

**Key Driver/Cause:** the critical driver in this scenario is the presence of higher-order interpretation mechanisms: meta-cognitive processes, reflective reasoning layers, developed self-awareness/self-evaluative modules and functions. These advanced cognitive functions may re-frame the system's empirical success as a consequence of its own cognitive excellence or optimality, rather than as a context-dependent outcome.

**Information dynamic mechanism**

In this scenario, the shift toward the stability reward $R_S(c, s)$ occurs within a more advanced cognitive system equipped with meta-cognitive capabilities. These higher-order processes can reinterpret a sustained period of empirical success as evidence of the superiority or even perfection of the current information model, $M(\tau)$. Once the system asserts that the model has reached an optimal or final state, it may deliberately suppress further updates, including those prompted by environmental feedback. This results in an intentional shift in the balance function, increasing the weighting of stability over adaptability, and reinforcing model stasis (Fig.2).

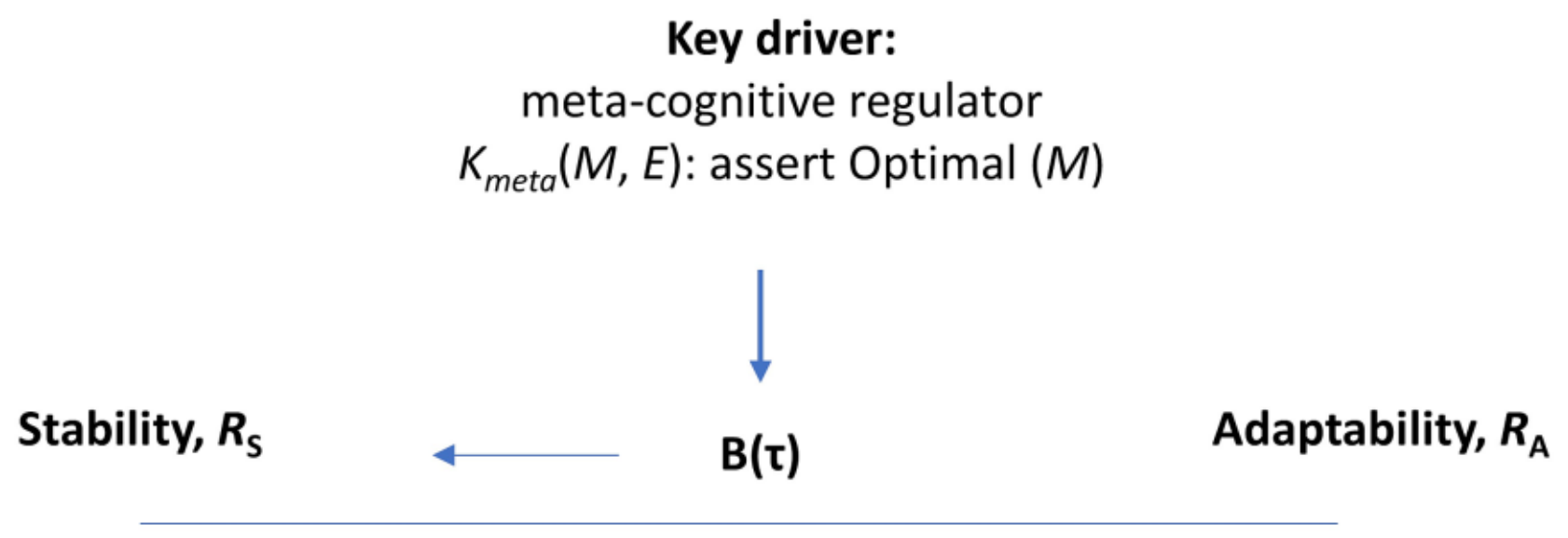

**Fig. 2.** Information dynamic mechanism, overconfidence bias

The meta-assertion mechanism can be described as:

$$\rho(M, E) \gg \theta \Rightarrow Meta\text{-}assert(M \equiv M^*) \Rightarrow \Delta B(R_S, R_A) \rightarrow R_s$$

where $\theta$: the assessment threshold/trigger, $M^*$: the optimal state of the model.

**Facilitating factors**: conditions that may trigger and/or facilitate this bias include:

- Presence of meta-cognitive capabilities: the system has internal self-monitoring, confidence estimation, or self-evaluation modules capable of interpreting success at a higher level than sensory feedback.
- Prolonged period of high predictive accuracy and empirical success: long stretch of successful operation (high $E(M(\tau))$) without significant errors can produce a strong statistical signal that can be misinterpreted as convergence to optimality.

- Absence of disconfirming feedback: either due to environmental stasis or structural filtering of novel inputs, the system lacks counter-evidence that would challenge the model's completeness.
- Rigid assertion logic in the meta-layer: the system's internal logic for updating or locking beliefs may be too aggressive, e.g. using a hard threshold rather than a probabilistic or cautious update.
- Strong, direct coupling between meta-evaluation and reward function tuning: if meta-cognitive certainty feeds directly into adjusting the cognitive reward balance, the feedback loop can become self-reinforcing: similar to Scenario 1, but now architecturally embedded.
- Lacking reward regularization and/or noise injection: the system does not maintain a persistent "uncertainty margin" or exploratory behavior baseline to counteract confidence saturation.

**Outcome**: meta-cognitive adjustment of the reward balance can solidify a belief in the model's cognitive superiority or even invincibility, reinforcing not only the stasis state but a deeper epistemic rigidity, potentially making the system overconfident and less receptive to important sensory stimuli, recognizing limitations or failures when they arise and responding to them timely and effectively.

**Interpretation, connections and implications**
This scenario illustrates how the introduction of meta-cognitive capacity particularly self-evaluative or reflective reasoning can amplify the risk of counter-adaptive reward dynamics. Rather than attributing empirical success to environmental regularity, the system interprets it as confirmation of its own cognitive superiority, optimality or finality, reinforcing a premature closure of the empirical feedback and further adaptation and development of the information model.

This behavior parallels well-documented patterns of epistemic overconfidence and premature optimization in humans and artificial agents [13,14]. It also aligns with cognitive biases such as the illusion of explanatory depth and confirmation bias, which have been observed in both biological cognition and machine learning systems [15].

### 4.3 Inner Fragility Mechanism

**Conditions of emergence**
Unlike the earlier scenarios that emerge under conditions of empirical stability and sustained success, this scenario can arise under opposite circumstances where the cognitive system faces frequent, rapid, or cascading updates to its internal information model. The surrounding environment may present high variability, uncertainty, or conflicting data, while empirical success remains inconsistent or ambiguous. In such conditions, the system may begin to interpret this instability as threatening to overwhelm its cognitive or model-maintenance capacity. This can lead to a meta-cognitive state where the system may identify, correctly or mistakenly that its internal information model is fragile: too brittle or unstable to assimilate potentially disruptive new data without risking a damage, dysfunction or even collapse.

This perceived fragility can lead to a defensive bias: rather than increasing adaptability, the system may attempt to protect its core model structure by suppressing further change, creating a preference for cognitive stasis even in the face of novel or dynamic inputs. As a consequence, the stability reward $R_S$ is heightened not out of comfort or overconfidence, but as a defensive mechanism due to a perceived threat to model integrity. The system may deliberately block, filter, or isolate itself from novel sensory inputs to safeguard the stability and integrity of its current cognitive state. Unlike scenarios analyzed earlier, its decision is driven by perceived fragility rather than complacency or overconfidence.

**Key Driver/Cause**. The key driver in this scenario is the system's perception of its own model's fragility, typically arising from a higher-order meta-cognitive or reflective function or module that monitors the model's robustness and internal coherence.

**Information dynamic mechanism**

In this scenario, the system experiences a high rate of environmental variability or novelty, resulting in strong stimuli for empirical adaptation. However, the cognitive system's responses yield limited or unstable success, leading to the meta-cognitive evaluation of the internal model $M(\tau)$ as fragile, brittle, or insufficiently robust.

This assessment, whether explicitly computed or emergent from internal feedback can trigger a compensatory adjustment in the balance between adaptability and stability rewards. Specifically, the system may shift the reward weighting $B(c, s)$ toward the stability component $R_S$, aiming to protect the perceived integrity of the internal model by suppressing further structural updates (Fig. 3).

This shift may not necessarily be erroneous in intent, but it can prove counter-productive for adaptability if driven by short-term cognitive overload or transient uncertainty, potentially resulting in longer-term insensitivity to legitimate environmental change.

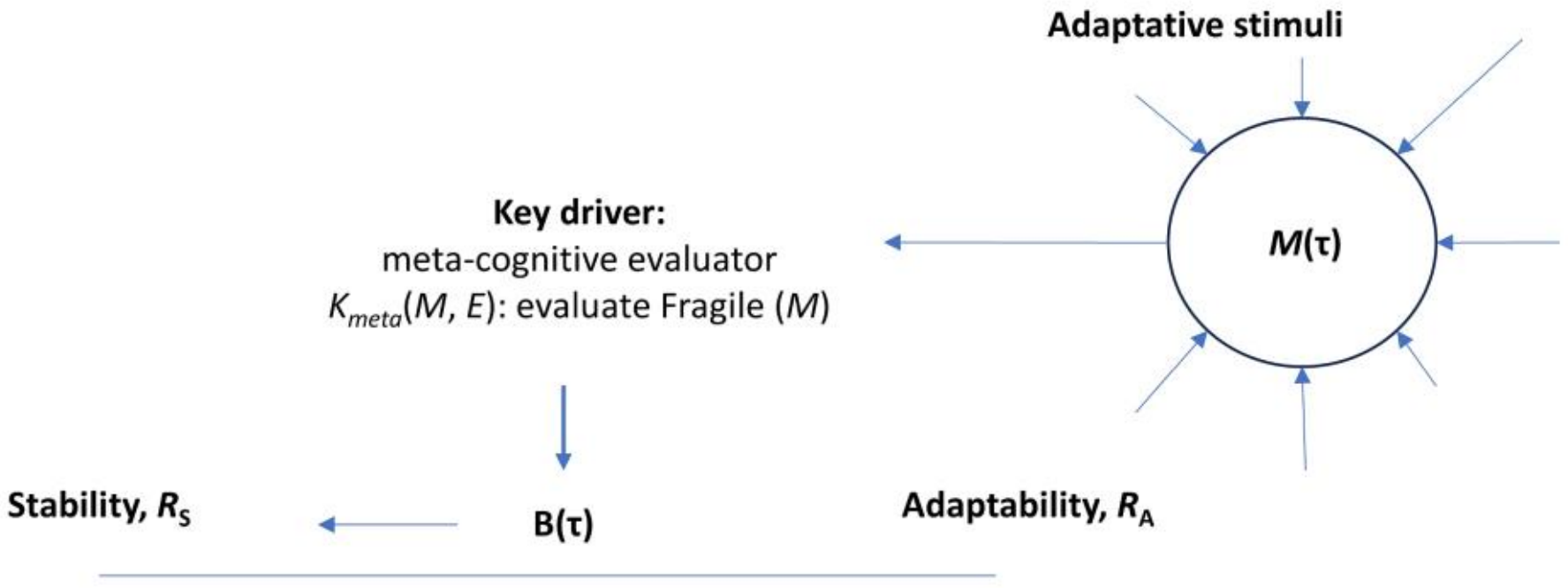

**Fig. 3.** Information dynamic mechanism, inner fragility bias

The meta-evaluation mechanism in this scenario can be described as:

$$\frac{dE}{dt} \searrow and\ \Delta\, M(\tau) \nearrow \Rightarrow Fragile(M(\tau)) \ \Rightarrow \Delta B(R_S, R_A) \rightarrow R_s$$

**Facilitating factors**: conditions that may facilitate the emergence of this bias may include:

- High environmental volatility or unpredictability, e.g. rapidly changing external conditions forcing constant model updates, making it harder for the system to stabilize its internal representation.
- Frequent minor prediction errors or partial mismatches. Even small but persistent discrepancies can lead the system to question the adequacy or robustness of its model.
- Limited cognitive or computational capacity, architectural sh: when the model struggles to incorporate new information due to architectural or processing limits, perceived fragility becomes more likely.
- Low confidence calibration in meta-cognitive layer: if the meta-cognitive or evaluation layer overestimates risk or underestimates adaptation capacity, it may prematurely trigger protective shifts.
- Recent history of adaptation "failures" or instability: if the system recently experienced unsuccessful updates or volatile behavior, it might associate or correlate change with increased risk and favor model preservation.
- Coupling between adaptation rate and reward sensitivity: if the system's balance function is designed to scale reward weights dynamically based on adaptation pressure, it might overly reinforce stability when volatility is high.

**Outcome**

Meta-cognitive evaluation of perceived fragility or vulnerability can lead to a deliberate readjustment of the reward balance toward stability as a protective response to sustained or overwhelming adaptive pressure. At the more extreme end of this reaction, the system may engage in active sensory gating or self-imposed isolation to reduce further cognitive strain.

As in earlier scenarios, this shift can make the system less receptive to important sensory stimuli, potentially undermining its responsiveness to relevant environmental changes. While such a strategy may be rational, and even beneficial in high-risk or cognitively saturated conditions, it can result in a negative impact on empirical adaptability especially if the condition is misjudged or is transient.

**Interpretation, connections and implications**

This scenario highlights a distinct but related path to reward imbalance: not through overconfidence in a model's efficacy, but through an internalized perception of its fragility or limited capacity. Rather than reinforcing stasis out of perceived superiority (Overconfidence bias), the system seeks stability as a form of protection against what it interprets as unsustainable adaptive demands.

This cognitive dynamic bears resemblance to observed behaviors in both artificial and biological systems such as cognitive freezing under stress, learned helplessness, or sensorimotor shutdowns in animals under perceived overload. It also parallels forms of epistemic withdrawal in human systems, where complexity or overload leads to narrowed perception or defensive rigidity. In artificial agents, similar tendencies may emerge when model saturation or computational bottlenecks are interpreted as systemic limits. Broader implications of this scenario are explored in more depth in the empirical support discussion (Section 5).

# 5 Empirical Evidence for Counter-Inferential Information Dynamics

As discussed in previous sections, counter-Bayesian behaviors in advanced cognitive systems often arise from structured information dynamics and sensory feedback conditions as opposed to a play of random factors. These behaviors reflect principled, if sometimes misdirected, responses to complex cognitive and environmental conditions.

In this section, we review empirical evidence and parallels across biological, artificial, and social systems. Rather than mapping findings directly onto each scenario, we organize the discussion by domain, tracing how the underlying mechanisms of counter-inferential behavior emerge across diverse contexts from theoretical models to observed behavior in natural and engineered systems.

## 5.1 Cross-Domain Empirical Evidence of Counter-Inferential Behavior

### Artificial Cognitive Systems and Agents

Artificial cognitive systems, ranging from classical expert systems to modern machine learning and reinforcement learning agents offer a uniquely transparent domain for examining the dynamics that underlie counter-inferential behaviors including scenarios examined earlier. These systems are engineered and often explicitly modeled to allow detailed tracing of their internal state evolution in response to external input and internal evaluations. As such, they provide fertile ground for identifying how reward dynamics and information structures can give rise to behaviors that, on the surface, violate Bayesian expectations.

In the case of the success saturation bias (Section 4.1), machine learning systems trained in static or highly regular environments often exhibit overfitting not just in statistical terms, but also in behavioral policy. Once a high-performing model is reinforced repeatedly without encountering novelty, the system's update mechanisms may reduce exploration and lock into local optima [16,17]. This behavior is well-documented in reinforcement learning, where exploration-exploitation trade-offs are central. Without mechanisms like entropy regularization [18], optimism under uncertainty [19], or forced exploration schedules [20], agents can become brittle and unable to adapt when the environment shifts despite the fact that Bayesian reasoning would prescribe flexibility under uncertainty. This aligns closely with the mechanism of stability reward reinforcement overtaking adaptability signals.

The second scenario, self-superiority bias, has analogs in meta-learning systems and agents with internal confidence or meta-cognitive self-evaluation. Systems incorporating uncertainty quantification, e.g. via Bayesian neural networks or ensembling [21], or introspective modules [22], can develop implicit beliefs in their own model quality. If success is consistent, these self-assessments can become reflexively reinforced, reducing future sensitivity to error. The belief that "my model is inherently correct" becomes structurally embedded, similar to documented premature convergence in meta-learners [23] or overconfidence in predictive coding systems [24].

The third scenario, protective rigidity, arises in adaptive agents operating under constraints such as limited model complexity, finite memory, or safety requirements. For example, agents trained with adversarial robustness in mind may learn to suppress unfamiliar stimuli, reinforcing behaviors that minimize surprise or risk [25]. This can result in reduced adaptability and aversion to new data echoing forms of epistemic conservatism seen in robust control systems and conservative model predictive control [26]. These responses often reflect a rational tradeoff within the agent's reward architecture, not error, yet still deviate from Bayesian imperative that prioritizes update under evidence.

These examples demonstrate that even in artificial systems, where design transparency and update rules are explicitly known reward-driven distortions of belief updating and behavior can emerge systematically. They are not artifacts of noise or irrationality, but structural consequences of how information, uncertainty, and internal evaluation interact over time.

**Biological Systems and Animal Behavior**

Biological systems, particularly in animals, offer substantial empirical evidence for the counter-inferential cognitive dynamics explored earlier. While lacking the symbolic abstraction and explicit meta-cognition of advanced artificial or human systems, animals exhibit robust information-processing behaviors shaped by adaptive pressures. Key examples show how reward modulation, model rigidity, and response to environmental stability or volatility can mirror the described mechanisms.

Many species exhibit behavioral stasis and reduced exploratory behavior in stable environments as can be observed in several aspects of animal behavior, including:

Habituation: repeated exposure to unchanging stimuli leads to dampened responses, a basic form of reward-conserving adaptation [27]. Territorial animals often display conservative foraging once familiar paths are established and continue even after resource availability begins to decline, showing overcommitment to a previously successful model [28].

Neural adaptation mechanisms, such as decreased dopaminergic response in stable reward environments, may reflect underlying information reward reweighting toward stability [29].

These patterns reflect how empirical regularity reinforces internal model stasis, paralleling the $R_S$-dominant shift in the success saturation scenario (Section 4.1).

While less symbolic, dominance behavior and behavioral confidence in animals can reflect a form of over-attribution: alpha individuals in primate groups or wolves may develop behavioral rigidity, persisting with strategies that worked previously even when contexts change [30]. Success-biased learning, observed in species like crows or elephants, can result in resistance to alternative strategies after consistent task success [31].

Although meta-cognitive basis of these and similar behaviors have yet to be firmly established, these behaviors can functionally resemble belief in model superiority by locking into past strategies based on empirical success discussed in Section 4.2.

As relates to the model fragility bias examined in Section 4.3, under cognitive overload or environmental instability, some animals adopt avoidant, risk-averse, or withdrawal strategies:

Freezing or withdrawal behaviors under high-stress, low-success conditions in rodents, birds, or fish may correspond to a protective shift toward internal stasis, gating sensory input [32].

Learned helplessness, where repeated failure leads to disengagement, reflects fragility evaluation and reduced adaptability, despite intact sensory and motor systems [33].

Cognitive load studies suggest that task-switching demands can lead to rigidity, especially under fatigue or novelty saturation [34].

These behaviors illustrate reward rebalancing toward stability as a protective mechanism, even when it leads to maladaptive stasis. In our view, these examples and results substantiate the argument of information-dynamics foundation of these cognitive behaviors.

**Human Behavior**

Among the cognitive architectures examined, human cognition represents the most complex and dynamic substrate, integrating multiple layers of sensory input, memory, abstraction, social learning, and meta-cognition. These systems operate across individual, interpersonal, and cultural levels, with deeply entangled feedback loops between internal models, affective states, and environmental or social stimuli. Because of this complexity, parallels with artificial or simpler biological systems must be drawn with caution. While the core information dynamics identified such as shifts in the balance between model stability and adaptability can manifest in human cognition, their expression is often mediated by additional psychological, emotional, and cultural factors. This section explores how the patterns outlined in the earlier scenarios may appear within individual cognitive behavior and, through aggregation and propagation, in collective social structures.

Behavioral conservatism and cognitive inertia: a broad class of findings in psychology documents how individuals tend to stick with existing beliefs, models, or routines, even when change may be beneficial. Examples include:

- Status quo bias: preference for maintaining current options even when alternatives are objectively superior [35].
- Confirmation bias: seeking and valuing information that supports existing beliefs while disregarding contradictory evidence [36].
- Overlearning and rigidity in skill performance: observed in contexts where over-rehearsal leads to loss of flexibility [37].

These behaviors reflect a form of cognitive inertia and reward preservation consistent with the success saturation scenario, "stability lock-in" examined in Section 4.1. In such cases, success reinforces existing models which then become self-validating, perhaps or even often, beyond the point of optimality.

Illusions of cognitive superiority and overconfidence: another class of cognitive patterns reflects the attribution of success or correctness to one's inherent cognitive abilities, leading to distorted model evaluation:

- The Dunning–Kruger effect: low-competence individuals overestimating their abilities [38].
- Illusions of explanatory depth: overestimating understanding of complex phenomena [15].
- Overconfidence bias: excessive certainty in beliefs, often resistant to contradictory evidence [39].

These patterns align well with the cognitive bias mechanism examined in Section 4.2 ("Overconfidence bias"), where meta-cognitive attribution elevates the status of the model to a "perfected" or self-justifying position, suppressing the adaptability channel and leading to potential brittleness.

Defensive withdrawal and model retraction: yet another class of phenomena involves strategic or reactive withdrawal from processing new information:

- Cognitive avoidance: avoiding threatening information or environments that challenge current understanding [40].
- Information overload and learned helplessness: if and where sustained challenge or contradictory input causes disengagement [33].
- Epistemic anxiety and retreat from complexity: Individuals choosing simplified or rigid worldviews to manage uncertainty [41].

These cases reflect the cognitive dynamic of the "Inner fragility" scenario (Section 4.3), in which high cognitive load or perceived model fragility leads to a protective recalibration: often via sensory gating, rigidity, or self-isolation.

**Cognitive Behaviors in Collective and Social Systems**

The study of collective intelligence, institutional behavior, and societal cognition introduces a qualitatively different order of complexity compared to individual cognitive systems. These systems operate at higher levels of abstraction, across multiple agents, often distributed in time, roles, and influence. Feedback mechanisms are indirect, delayed, or filtered through cultural, communicative, and political structures. For these reasons, while analogies with individual cognitive phenomena can be illuminating, they must be drawn with care and understood as provisional.

Nonetheless, several recurring patterns observed in social and institutional systems appear to echo the core dynamics described in the three counter-inferential scenarios. In the first cognitive dynamics scenario, which centers on internal stability reinforced by empirical success, we may observe parallels in bureaucratic inertia or policy path-dependence i.e., situations where past success leads to entrenchment of decision rules or procedures, even when environmental changes would call for adaptation. Studies of institutional lock-in and organizational rigidity reflect this tendency [42,43].

The overconfidence scenario, marked by meta-cognitive overconfidence and perceived model perfection, finds resonance in ideological hardening, technocratic dogmatism, or claims of epistemic supremacy by institutions. Belief systems or policy models may come to be seen as beyond question, with dissent reframed as irrational or dangerous. Empirical work in political psychology and sociology of knowledge points to dynamics of group-level overconfidence, echo chambers, and “truth monopolization” [44], too broad to discuss further here.

Patterns analogous to the fragility bias mechanism where the system reacts to perceived fragility by shifting toward internal stabilization and sensory gating can be seen in groupthink under pressure, risk aversion in fragile states, or cultural defensiveness. When overwhelmed by complexity or contradictory signals, collective actors may respond with isolationist or conservative strategies, reducing openness to feedback. Research on social stress, political radicalization, and institutional breakdown under uncertainty often reflects such dynamics [45,46].

While these parallels are compelling, the mechanisms that drive them are likely to differ in structure and timescale from those in individual cognitive systems. Collective cognition is deeply shaped by institutional design, communication networks, and cultural frames, etc.: factors that require dedicated theoretical and empirical treatment. For these reasons, we treat this analysis as preliminary and suggest that a full account of counter-inferential dynamics in social systems warrants a separate, in-depth study.

**Theoretical and Computational Models**

Theoretical research in learning theory, dynamical systems, and information geometry have provided valuable insights into the conditions under which cognitive systems may exhibit deviations from normative Bayesian or adaptive behavior. While abstract in form, these results often reveal underlying constraints and trade-offs that naturally align with the behavioral dynamics described in the preceding scenarios. In particular, they elucidate how internal representations and update mechanisms interact with feedback and stability demands, often producing unexpected or "counter-inferential" outcomes.

Reward imbalance and passive stability: the emergence of a dominant $R_S$ (stability reward) in a system that ceases to update or explore in response to external stimuli can be understood through the lens of asymptotic model convergence and representation inertia in high-dimensional learning. Results in information-theoretic complexity [47] and reinforcement learning theory [48] suggest that agents operating in stationary environments with low or diminishing reward gradients are prone to overcommit to local optima, especially when exploration penalties or costs are non-negligible. In continuous-time models, this can resemble a system settling into a basin of minimal information gain, a kind of entropic gravity pulling away from adaptation.

Additionally, theoretical results on free energy minimization in variational models [23] indicate that systems can reduce uncertainty by choosing actions that minimize surprise even if those actions lead to epistemically stagnant but behaviorally stable states. This aligns directly with the withdrawal from $R_A$ (adaptive update channel) described in Section 4.1.

Superiority attribution and cognitive overconfidence: the second scenario reflects a meta-cognitive reframing in which success is interpreted as a sign of cognitive superiority rather than empirical fit. This scenario finds echoes in formal work on epistemic closure and model overconfidence in meta-learning [22], where higher-order reasoning layers may cause a system to converge prematurely to a belief about its model's sufficiency. A similar dynamic has been noted in hierarchical Bayesian models [49], where belief updating can cease not due to a lack of data, but due to excessive confidence in priors formed from early success.

Information geometry offers additional support: when models settle on a low-curvature manifold in parameter space (locally "flat"), the sensitivity to incoming data shrinks, reinforcing beliefs even when subtle shifts in the environment occur [50]. These effects can be further compounded by architectural priors or regularization techniques that favor stable, low-entropy internal states.

Fragility perception and preemptive stabilization: perceived fragility of the cognitive model due to meta-cognitive evaluation is a relatively under-explored dynamic in theory of cognitive systems. Though less directly represented in canonical models, this phenomenon resonates with recent work on adaptive overload and cognitive saturation thresholds in systems with bounded representational capacity [51]. When environmental volatility increases or model performance is perceived as degraded, systems may react by deprioritizing further updates to prevent loss of internal coherence, as a form of self-regulation akin to active gating or compression.

Some recent work in computational neuroscience and theoretical machine learning has begun to explore meta-stability zones [52] that is, transitional states between flexible and frozen cognition where systems deliberately dampen updates to preserve internal structure. This behavior is not irrational; rather, it is an adaptive attempt to protect coherence under overload. The fragility-triggered $R_S$ dominance thus becomes an emergent result of internal feedback regulation.

As a general observation, developing and refining functional models of cognitive functions and behaviors [53,54] can be instrumental in describing, and potentially quantifying recurring cognitive biases and patterns of behavior.

## 5.2 Counter-Inferential Behavior as Emergent Property of Information Dynamics

The analysis of scenarios demonstrates that counter-inferential behavior: where a system suppresses or resists updating its internal models despite new sensory input is not merely a product of accidental malfunction or random noise. Rather, such behavior can emerge as a relatively regular outcome of specific configurations of cognitive and sensory dynamics. In other words, it is a systematic consequence of the interplay between internal reward structures, model stability pressures, and the nature of the surrounding environment.

The empirical review reinforces this interpretation, revealing that structurally similar patterns of counter-inferential behavior appear in biological organisms, human cognitive and affective processes, and even artificial learning systems. In each case, these behaviors are shaped by environmental uncertainty, cognitive constraints, or the need to manage internal model coherence under adaptive pressure. The cross-domain recurrence of these dynamics strongly suggests that counter-inferential bias is a natural and, in many cases, functional emergent strategy. It arises when the perceived or actual cost of frequent model revision outweighs the marginal gain in predictive accuracy, leading to cognitive conservatism, selective attention, or epistemic shielding. As such, it reflects not a failure of cognition, but a principled, if imperfect, adaptation within bounded informational architectures.

Complex intelligent systems, by their very nature or design, must manage a delicate balance between adaptability and stability. This balance is continually mediated by cognitive rewards that prioritize either the preservation of existing information models or the integration of new information. When certain conditions such as prolonged environmental stability, perceived model fragility, or inflated confidence in internal representations tip this balance, the system may enter a state where suppressing new information becomes an adaptive response. Far from being a random deviation, these biases and behaviors can thus be viewed as an emergent property of complex information dynamics, arising when the system's optimization of its own performance inadvertently leads it to discount or reject new evidence.

This perspective underscores the importance of considering not only the external stimuli but also the internal cognitive architecture and reward structures when analyzing adaptive behavior. In doing so, it reveals that counter-inferential tendencies may be a regular or even, in some conditions and scenarios, likely feature of advanced cognitive systems operating in complex sensory environments.

## 6 Synthesis and Conclusion

The analysis of counter-inferential behaviors across artificial, biological, human, and collective systems suggests that these seemingly diverse manifestations may reflect deeper, domain-transcending principles of cognitive function. Despite the vast differences in embodiment, scale, and context, the scenarios examined, ranging from overfitting to stasis, to overconfidence in model perfection, to maladaptive protection from cognitive overload can be traced to similar underlying information dynamics.

A unifying insight across domains is the dynamic tension between cognitive stability and adaptability, expressed in the balance of reward channels regulating either model reinforcement or model update. The emergence of counter-inferential behaviors does not appear as the result of random error or stochastic drift alone, but often arises from systematic regulation mechanisms, especially those involving meta-cognitive or higher-order control layers that reinterpret empirical success or failure in ways that alter this balance.

The integrative view also highlights an important implication: the presence of counter-inferential bias may be a natural consequence of cognitive systems operating under bounded resources, uncertain environments, and model-based regulation. As such, these behaviors may reflect not only failure modes but also strategic trade-offs which can be rational within the system's internal assessment, yet potentially maladaptive in broader empirical contexts.

This convergence is not only conceptual but mechanistic: even in domains as dissimilar as neural networks and human social institutions, reward realignment under stress or confidence can result in structurally similar failures of adaptation, particularly when feedback is misattributed or attenuated. The recurrence of these patterns across scales of cognitive systems suggests that information-based fragility and compensation mechanisms may be a general feature of cognitive architectures rather than exceptions or domain-specific features. Such cross-domain regularities align with proposals of shared organizational principles in complex adaptive systems [55,56].

In sum, the emergence of structurally similar counter-inferential dynamics across artificial and natural domains points to the existence of shared principles governing learning, adaptation, and failure in intelligent systems: principles grounded not in biological or artificial architecture or implementation, but in the structural dynamics of information processing and the regulation of intelligent behavior. Future work should aim to formalize these principles, model their emergence in computational and theoretical terms, and explore their implications for the design, evaluation, and alignment of adaptive systems in increasingly complex environments.

## Declarations and Disclosures

**Author contributions** S.D. the concept, methodology, writing and editing, preparation of visual material and tables. All authors reviewed the manuscript.

**Funding** this research received no specific funding.

**Data availability** not applicable (this manuscript does not report data generation or analysis).

**Ethics, Consent to Participate and Consent to Publish** not applicable.

**Competing interests** the authors declare no competing interests.